\documentclass[10pt,twocolumn,letterpaper]{article}

\usepackage{paperstyle}

\usepackage{cuted} %

\usepackage{currfile} %

\usepackage{caption} %

\usepackage{multirow}
\usepackage{placeins} %

\usepackage{microtype}

\definecolor{linkblue}{rgb}{0.21,0.49,0.74}
\usepackage[pagebackref,breaklinks,colorlinks,allcolors=linkblue]{hyperref}

\title{ProxyVerse-15K: Scaling Editable 3D Proxies for Real Video\\through Harness Distillation}

\newcommand{\projecturlraw}{https://garfield-kh.github.io/ProxyVerse/}
\newcommand{\projecturl}{\expandafter\url\expandafter{\projecturlraw}}
\newcommand*{\acknowledgmentstext}{}
\newcommand{\suppname}{appendix}
\author{%
Kehong Gong$^{1}$\quad
Guanli Hou$^{2}$\quad
Xiaoxu Zheng$^{1}$\quad
Seunghun Lee$^{1}$\quad
Mingxi Xu$^{2}$\quad
Zhengyu Li$^{2}$\\
Daisheng Jin$^{2}$\quad
Yichen Zhou$^{1}$\quad
Weijia Wu$^{2}$\quad
Xiaoyu He$^{2}$\quad
Mingyuan Zhang$^{2,*}$\quad
Michael Bi Mi$^{1}$\\[0.6ex]
{\normalsize $^{1}$Huawei Technologies Co., Ltd\qquad $^{2}$Huawei Noah Research Institute\qquad $^{*}$Corresponding author}
}

\begin{document}
\maketitle
\begin{strip}
\centering
\includegraphics[width=\linewidth]{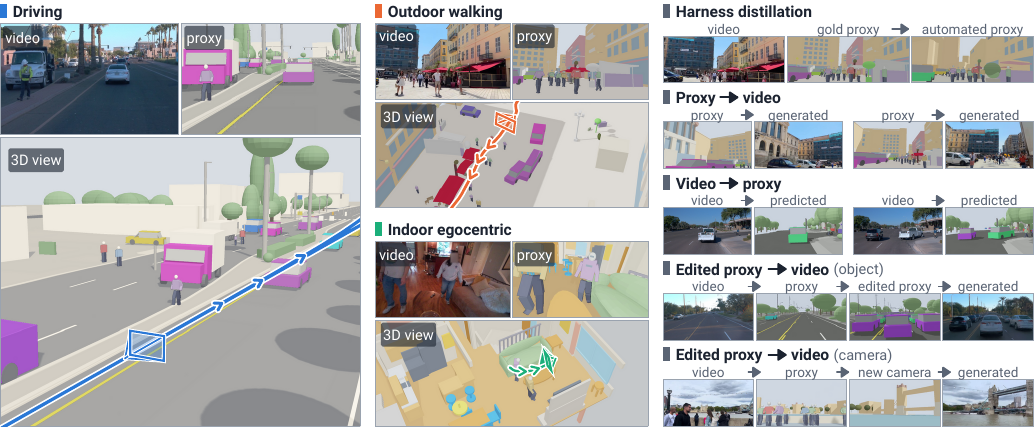}
\captionsetup{hypcap=false}
\captionof{figure}{ProxyVerse pairs real videos with frame-aligned 3D proxies in driving, outdoor walking and indoor egocentric video. A proxy is a compact 4D scene of primitives, templates and a camera path; it renders from the recorded camera or any other view and is editable object by object. Right, top to bottom: harness distillation carries gold-proxy quality to automated proxies; proxies drive video generation; the new video-to-proxy task predicts a proxy from video; object and camera edits of a proxy are rendered back into video, shown next to the original video.}
\captionsetup{hypcap=true}
\label{fig:teaser}%
\end{strip}

\begin{abstract}
We introduce ProxyVerse-15K, which pairs 14{,}923 real-video clips (112.0 hours) across driving, outdoor walking and indoor egocentric domains with frame-aligned proxies, compact 4D scenes of typed primitives, posed object and human templates, and a camera path, renderable from new views and editable object by object.
To build them at scale, we propose \emph{harness distillation}: on a gold subset, whose proxies are verified against the video, a strong multimodal model performs human-guided recursive self-improvement through a render--compare--diagnose--edit--verify loop, and generalizable improvements are consolidated into a harness of detectors, repairs, narrow visual questions, layer modules and templates that open models run on the remaining clips without task-specific training.
In proxy-to-video, removing proxy control reduces PSNR by 3.5--5.2\,dB across the three domains, and with the same proxy our multi-domain baseline has higher PSNR than a much larger commercial video model in every domain.
For the new inverse \emph{video-to-proxy} task, a feed-forward baseline predicts typed 3D objects and trajectories, ground and static structure from posed video; trained on automated proxies and evaluated against gold, it improves driving and outdoor object prediction over a released monocular 3D detector.
Proxy edits transfer object, motion and camera changes to the generated video.
We will release the videos, proxies, renderings, harness code and baselines.
\ Project page: \href{\projecturlraw}{\upshape garfield-kh.github.io/ProxyVerse}\end{abstract}

\section{Introduction}
\label{sec:intro}
\providecommand{\cmark}{\checkmark}
\providecommand{\xmark}{$\times$}
\providecommand{\nomark}{--}
\providecommand{\pmark}{$\circ$}
\begin{table*}[t]
\caption{ProxyVerse and the data used for proxy-, box- and layout-conditioned video generation, as reported by the cited sources. \emph{Static} and \emph{Dynamic}: object-level 3D for static structure and for moving objects; \emph{Articul.\ people}: people carry a body pose. \pmark: partial; \nomark: not stated.
$^{a}$Road-map layers.
$^{b}$Depth, poses and masks, no object labels.
$^{c}$Detected 3D boxes for gameplay recordings.
$^{d}$Skeleton variant.
$^{e}$Ground plane only.
$^{f}$Mesh lifted from first-frame depth.
$^{g}$Driving boxes and indoor body poses come from the sources.
$^{h}$Released with this paper.}
\label{tab:compare}
\centering
\footnotesize
\setlength{\tabcolsep}{3.2pt}
\begin{tabular}{@{}l l l l c c c l c@{}}
\toprule
 & Video & & & Static & Dynamic & Articul. & Control & \\
Dataset (used by) & source & Domain & Scale & in 3D & in 3D & people & produced by & Public \\
\midrule
\multicolumn{9}{@{}l}{\textit{Synthetic, game or generated video}} \\
360\textdegree-Motion \cite{fu20253dtrajmaster} & Unreal Engine & humans, animals & 54k videos & \xmark & \cmark & \xmark & engine render & \cmark \\
SynFMC \cite{shuai2025freeform} & Unreal Engine & open & 62k videos & \xmark & \cmark & \xmark & engine render & \cmark \\
OmniWorld-Game \cite{zhou2026omniworld} & games & games & $>$96k clips, $>$214\,h & \xmark$^{b}$ & \xmark$^{b}$ & \xmark & game engine & \cmark \\
Coarse-to-Real \cite{gomeznogales2026coarsetoreal} & 3D tools & urban crowds & 1.3k paired clips & \cmark & \cmark & \cmark & engine render & \nomark \\
CoCode-World \cite{li2026cocodeworld} & GTA V & game, urban & 7{,}846 clips & \cmark & \cmark & \pmark$^{d}$ & engine state & \nomark \\
AnyGameRenderer \cite{liu2026anygamerenderer} & games, UE & games & 10{,}330 clips, 14.8\,h & \pmark$^{c}$ & \cmark & \xmark & render; 3D detection & \nomark \\
CoaG \cite{yang2026coag} & T2V model & people & 1{,}935 clips & \pmark$^{e}$ & \cmark & \xmark & lifted from video & \cmark \\
\midrule
\multicolumn{9}{@{}l}{\textit{Real video}} \\
nuScenes \cite{caesar2020nuscenes} (\cite{gao2024magicdrive,wen2024panacea}) & real & driving & 1k scenes, 5.5\,h & \pmark$^{a}$ & \cmark & \xmark & human annotation & \cmark \\
RDS-HQ (\cite{nvidia2025cosmostransfer1,ren2025cosmosdrivedreams}) & real & driving & 65k clips, $\sim$360\,h & \pmark$^{a}$ & \cmark & \xmark & auto-label + QA & \nomark \\
CineMaster data \cite{wang2025cinemaster} & real & open & 156k videos & \xmark & \cmark & \xmark & estimated & \nomark \\
LooseControlVideo data \cite{bhat2026loosecontrolvideo} & real & open & $\sim$10k videos & \xmark & \cmark & \xmark & estimated & \nomark \\
Mise-en-Video data \cite{zou2026miseenvideo} & real & film, TV & 160k videos & \xmark & \cmark & \xmark & estimated & \nomark \\
LiVERSet \cite{cai2026lighting} & real, synthetic & scenes, objects & $>$11k clips & \pmark$^{f}$ & \xmark & \xmark & estimated; render & \nomark \\
\midrule
ProxyVerse (ours) & real & driving, walk, indoor & 14{,}923 clips, 112.0\,h & \cmark & \cmark & \cmark & automatic$^{g}$ & \cmark$^{h}$ \\
\bottomrule
\end{tabular}
\end{table*}

Video models synthesize realistic footage, but controllable generation requires explicit specification of scene content, object motion and camera motion.
Text is imprecise, while depth, edge and segmentation sequences are cumbersome to edit consistently~\cite{zhang2023adding,jiang2025vace,nvidia2025cosmostransfer1}.
We study a \emph{3D proxy} as an interface that exposes persistent objects, static structure, articulated people and camera motion in one editable scene: a compact, untextured 4D scene of typed primitives, posed object and human templates, and a camera path.
It can be edited object by object and rendered from any view, fixing structure and motion while leaving appearance to the generator.
Paired video--proxy data therefore supports both generating video from an edited proxy and recovering an editable proxy from video.

Learning both video-to-proxy and proxy-to-video requires paired data.
Existing complete 3D controls come mainly from rendered worlds~\cite{gomeznogales2026coarsetoreal,li2026cocodeworld,liu2026anygamerenderer}; real-video controls usually cover only moving boxes~\cite{wang2025cinemaster,bhat2026loosecontrolvideo,zou2026miseenvideo} or driving boxes on road maps~\cite{caesar2020nuscenes,nvidia2025cosmostransfer1} (\cref{tab:compare}).
None combines real video, static structure, moving objects and articulated people in one multi-domain representation.
The missing resource is thus scalable supervision connecting real video to editable scene-level 4D structure.

Constructing such pairs is difficult: noisy frame-wise perception must become a persistent, internally consistent scene.
We introduce \emph{harness distillation}, which spends an expensive multimodal model and human review on improving the reusable construction system rather than on finishing every proxy.
On a gold subset, a strong multimodal model repeatedly renders, compares, diagnoses, edits and verifies proxies, with targeted human feedback.
Generalizable improvements become detectors, repairs, narrow visual questions, layer modules and templates, forming a harness that open models execute over the remaining clips without training during dataset construction.

\begin{figure*}[t]
\centering
\includegraphics[width=\linewidth]{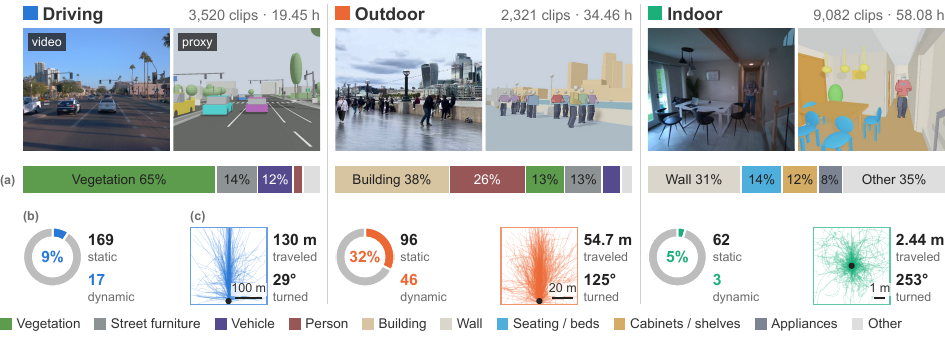}
\caption{ProxyVerse at a glance. Top: one frame per domain with the proxy rendered from the same camera. (a) Share of proxy instances per category, colored as in the renderings (counting rules differ by domain; indoor counts include walls, doors and windows). (b) Median static and dynamic objects per clip (per window indoors, objects in view); dynamic share = median dynamic count over the sum of the two medians. (c) Camera paths of 500 random clips per domain, aligned at their start, with the median path length and total turning per clip: driving cameras travel more than twice as far as walking ones and indoor cameras stay within a few meters.}
\label{fig:stats}
\end{figure*}

Using this process, we construct \emph{ProxyVerse-15K} (ProxyVerse below), containing 14{,}923 real-video clips, 112.0 hours and 5.66 million frames across driving, outdoor walking and indoor egocentric video.
Each clip has a temporally aligned proxy with cameras, structure, ground, typed objects and articulated people when applicable (\cref{fig:teaser,fig:stats}); gold proxies define the evaluation sets.
The remaining proxies are produced by the distilled harness with open models.

ProxyVerse supports proxy-to-video control and the new inverse \emph{video-to-proxy} task.
Our feed-forward baseline predicts objects and trajectories, ground and static bird's-eye-view (BEV) structure, which rules and templates assemble into a proxy.
Experiments provide three forms of evidence.
First, the automated harness raises agreement with the gold proxies on the outdoor test clips from a person F1 of 0.372 for perception initialization to 0.815.
Second, removing proxy control reduces generation PSNR by 3.5--5.2\,dB across the three domains, and with the same proxy the multi-domain baseline has higher PSNR than a much larger commercial video model in all three domains.
Third, trained on automated proxies and evaluated against gold, the video-to-proxy baseline improves object prediction over a released monocular 3D detector while predicting all major proxy layers.
Editing objects, motion and camera paths further produces corresponding changes in generated video.

Our contributions are:
\begin{itemize}
    \item \textbf{ProxyVerse-15K}, an open dataset pairing 14{,}923 real-video clips across three domains with temporally aligned, editable 3D proxies, establishing paired supervision between video and scene-level 4D structure.
    \item \textbf{Harness distillation}, which consolidates improvements accumulated through human-guided recursive self-improvement into a reusable harness of detectors, repairs, narrow visual questions, layer modules and templates, enabling open models to construct proxies at scale.
    \item \textbf{The new video-to-proxy task and a feed-forward baseline}, which jointly predicts the major components of an editable proxy from posed video. With gold-proxy test sets and baselines for both directions, it supports proxies as a bidirectional interface for structured video understanding and controllable generation.
\end{itemize}

\section{Related work}
\label{sec:related}

\paragraph{Controllable video generation and 3D proxies.}
Video diffusion models~\cite{blattmann2023stable,yang2025cogvideox,kong2024hunyuanvideo,wan2025wan} are controlled by branches that read depth or edges, following ControlNet~\cite{zhang2023adding} (for video, VACE~\cite{jiang2025vace} and Cosmos-Transfer~\cite{nvidia2025cosmostransfer1}), by camera embeddings~\cite{wang2024motionctrl,he2025cameractrl}, lifted geometry~\cite{ren2025gen3c,yu2025trajectorycrafter,cao2025uni3c,gu2025diffusion}, and driving layouts~\cite{gao2024magicdrive,gao2025magicdrivev2,wen2024panacea,wang2024drivedreamer}.
Closer to proxies, generators use 3D boxes~\cite{bhat2024loosecontrol,eldesokey2025build}, boxes or 6DoF trajectories~\cite{wang2025cinemaster,fu20253dtrajmaster}, sparse oriented blocking boxes~\cite{bhat2026loosecontrolvideo}, or a ground grid with one cylinder per person~\cite{yang2026coag}.
Lacking real paired data, CoaG constructs pairs from generated clips, Coarse-to-Real~\cite{gomeznogales2026coarsetoreal} uses a small synthetic set, VideoFrom3D~\cite{kim2025videofrom3d} avoids paired data and ProxyUp~\cite{zhou2026proxyup} is training-free.
ProxyVerse instead pairs real videos with proxies containing static structure, objects and people across three domains.

\paragraph{Datasets with 3D annotation, and recovering scenes from video.}
Driving datasets pair real video with road-user boxes~\cite{caesar2020nuscenes,sun2020scalability}, while indoor datasets provide reconstructions, boxes or aligned CAD models~\cite{dai2017scannet,baruch2021arkitscenes,avetisyan2019scan2cad,lazarow2025cubify}.
Aria Digital Twin~\cite{pan2023aria} aligns 200 egocentric sequences in two rooms with a digital twin; KITTI-360~\cite{liao2021kitti360} labels urban scenes with 3D primitives, and CAD-Estate~\cite{maninis2023cadestate} places CAD models in real-estate videos (full comparison in \cref{tab:compare_real}).
ProxyVerse combines static structure, dynamic entities and camera paths in one renderable representation across three domains.
For recovery, feed-forward reconstruction predicts cameras, depth and points~\cite{wang2024dust3r,wang2025vggt,lin2026depth} but not typed objects.
Layout estimators~\cite{zou2018layoutnet,sun2019horizonnet}, scene-language models~\cite{avetisyan2024scenescript,mao2025spatiallm} and 3D detectors~\cite{brazil2023omni3d,zhang2025detect,detone2026boxer,straub2024efm3d} each recover part of a proxy; we compare with DetAny3D in \cref{sec:exp:v2p} and with SpatialLM and Boxer in the \suppname{} (\cref{tab:v2p_indoor}).

\paragraph{Language-model agents for 3D scenes, and reusing their experience.}
Language and vision-language models generate 3D scenes by writing layouts or code and inspecting renders~\cite{feng2023layoutgpt,yang2024holodeck,hu2024scenecraft,sun20253dgpt,huang2024blenderalchemy,sun2025layoutvlm}; our gold stage instead works against an existing video, so edits can be checked by overlay.
Agent experience persists without weight updates as skill libraries~\cite{wang2024voyager}, induced workflows~\cite{wang2025agent} or extracted rules~\cite{zhao2024expel,chen2024automanual}; in data labeling, programs replace per-item queries~\cite{ratner2017snorkel,huang2024alchemist}.
Harness distillation applies this principle to dataset construction, checking detectors, repairs and questions on gold proxies before open models run them at scale.

\section{The ProxyVerse dataset}
\label{sec:dataset}

ProxyVerse pairs real videos with frame-aligned 3D proxies in driving, outdoor walking and indoor egocentric domains.
Each clip includes the source video, proxy scene, per-frame camera poses and intrinsics, and proxy rendering.
In total it has 14{,}923 clips, 112.0 hours and 5.66 million frames: 19.45 hours of driving, 34.46 hours outdoors and 58.08 hours indoors (per-source counts in \cref{tab:scale}).

\paragraph{Domains and sources.}
The driving domain has 3{,}520 clips from PhysicalAI-AV~\cite{nvidia2025physicalai}, Waymo~\cite{sun2020scalability} and nuScenes~\cite{caesar2020nuscenes}, with a median length of 20 seconds; cameras and road-user boxes come from the source datasets, with ground snapping and template-size adjustment.
The outdoor domain has 2{,}321 roughly one-minute clips from 219 walking videos in Sekai~\cite{li2025sekai} and MiraData~\cite{ju2024miradata};
camera poses are estimated from video and metric scale per clip from monocular priors, checked against people and car sizes, camera height and walking speed; clips with contradictory evidence are rejected.
The indoor domain has 9{,}082 24-second windows from 757 Nymeria~\cite{ma2024nymeria} recordings of 201 wearers in 45 homes; each home is modeled once, and people and held objects are added per recording from motion capture.

\paragraph{Proxy representation.}
\label{sec:dataset:repr}
A proxy is a list of categorized, oriented primitives and per-frame poses for moving objects; people are articulated templates driven by estimated or captured body pose.
It is rendered in flat category-coded colors (\cref{fig:stats}).
Appearance and surface detail are left to the video generator.

\paragraph{Splits and gold proxies.}
Driving has 3{,}197 training and 185 test clips, split by scene.
Indoor has 8{,}160 training and 866 test windows; the main tables use the 309 test windows from four homes unseen in training.
Outdoor has 2{,}210 training and 60 test clips; same-video train/test clips are at least two minutes apart. All test clips have gold proxies.
Gold proxies, built by a strong multimodal model and reviewed frame by frame, cover 185 driving clips, 142 outdoor clips and 866 indoor windows; the automated pipeline of \cref{sec:pipeline} produces the rest.

\section{Harness distillation and video-to-proxy}
\label{sec:pipeline}

\begin{figure*}[t]
\centering
\includegraphics[width=\linewidth]{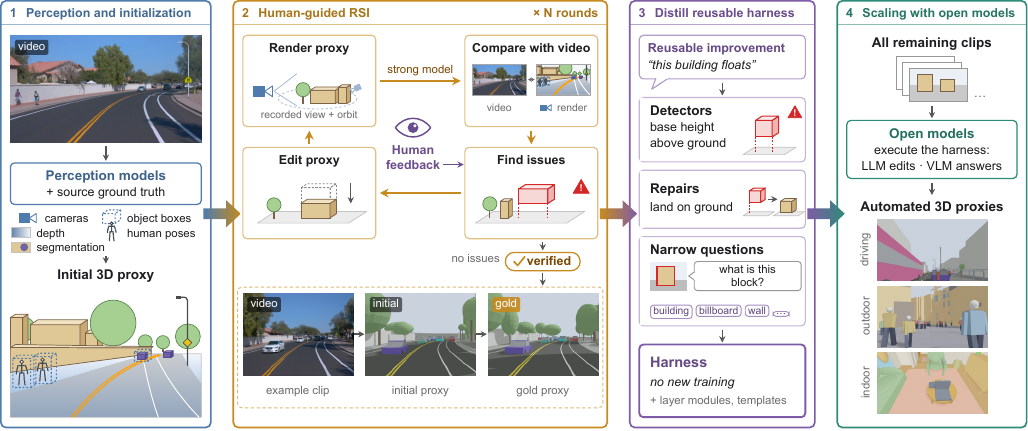}
\caption{Harness distillation. Perception and sensor data give an initial scene (\cref{sec:pipeline:init}). On a gold subset, a strong multimodal model repeatedly renders, compares, diagnoses, edits and verifies proxies, with targeted human feedback when it misses an issue (\cref{sec:pipeline:gold}). The proxy is rendered from the recorded camera for comparison with the video and from other views for inspection. Generalizable improvements are consolidated into detectors, repairs, narrow questions, layer modules and templates (\cref{sec:pipeline:hd}). Open models then execute the improved harness at scale (\cref{sec:pipeline:scale}). Bottom of stage 2: a later frame of the same clip, its proxy before human-guided refinement and its verified gold proxy. Indoors, the open models edit a home model assembled from the gold proxies of that home.}
\label{fig:pipeline}
\end{figure*}

A proxy must agree with every video frame while remaining internally consistent: elements have typed extents, and moving objects preserve identity.
Perception outputs are partial, noisy and often frame-wise, so converting them into persistent objects, trajectories, ground and static structure requires repeated decisions based on rendered comparisons with the video.
A strong multimodal model can make these decisions inside a tool loop, but running it with human review on every clip does not scale.
We therefore use human-guided recursive self-improvement (RSI) on a subset to improve a reusable harness that open models execute at scale (\cref{fig:pipeline}).

\subsection{Human-guided recursive self-improvement}
\label{sec:pipeline:gold}
\paragraph{Initialization.}\label{sec:pipeline:init}
The first step collects evidence and initializes the scene, using source ground truth when available.
In driving, cameras, LiDAR, road-user boxes and, where available, HD maps come from the datasets; fused depth, segmentation and fitted primitives supply static layout.
Indoors, device poses and point clouds come from the capture device and body motion from the motion-capture suit.
Outdoors, cameras, depth, instances and body poses are estimated with VGGT-Omega, Depth Anything~3~\cite{lin2026depth}, Pi3X~\cite{wang2026pi3,wang2025pi3x}, SAM~3~\cite{carion2026sam} and Human3R (full list in \cref{sec:suppl:percep}).

For a subset in each domain, a strong multimodal model, acting as a coding agent, improves both the proxy and its construction process inside a tool harness.
It starts from the same perception initialization as the automated pipeline.
The harness provides tools to edit the scene, render it from the real camera or an orbit, overlay it on the video and compute alignment metrics.
The agent follows a render--compare--diagnose--edit--verify loop; targeted feedback from people and Claude Opus re-enters the loop to correct missed issues.
An iteration can improve the current proxy, the reusable harness or both, and the resulting gold proxies define the quality target.

\subsection{Distilling the harness}
\label{sec:pipeline:hd}
When the RSI loop exposes a recurring problem, such as a floating building or a person disappearing for several frames, we ask whether its diagnosis and repair generalize beyond the current clip.
Generalizable improvements are consolidated into reusable components.
Where a defect can be measured from geometry or rendering, we write a detector that finds it in any clip and check its hits on gold proxies.
Each detector is paired with a local edit or a module that rebuilds a layer such as ground or buildings; the repair is re-rendered, checked and reverted if worse.

Some defects require the image, but open vision-language models (VLMs) answer a general better-or-worse question unreliably (\cref{sec:suppl:general_q}).
We therefore split reviewer judgment into narrow questions with a few answers, measure each question on confirmed examples and attach a fixed repair to each answer.
Detectors, repairs, questions, layer modules and templates form a harness with no newly trained weights.
The PhysicalAI-AV harness chains 81 stages; over 1{,}843 clips its detectors fired 33{,}739 times and 86.3\% of the fixes passed the re-check. The outdoor harness kept 4{,}645 fixes over the 2{,}261 outdoor clips with automated proxies.

\subsection{Scaling with open models}
\label{sec:pipeline:scale}
Once improvements are consolidated, an open language model builds and edits scenes and chooses tools, while an open VLM answers narrow questions; in driving and outdoors these are DeepSeek V4.1-Flash~\cite{deepseekai2026deepseekv4} and Qwen3.8-Flash-Next.
In driving and outdoor video, gold proxies are used to build and calibrate the harness, and the automated pipeline runs without gold at production time.
Indoors, the home model shared by all recordings of a home is assembled from the gold proxies of that home.
The domains share the harness abstraction and verification loop but instantiate it with domain-specific evidence, modules and templates.
Driving relies mainly on sensor-derived layers and annotated boxes; indoors, the model edits a shared home model for each recording; outdoors, it drafts a scene that rules and questions then correct.
An automated clip costs 1.6--2.0\,CNY of API calls in PhysicalAI-AV and takes 15--22 minutes outdoors, while a gold pass by the strong multimodal model takes 35--42 minutes per driving clip.
Starting from the same perception initialization as the gold stage, the open-model harness raises agreement with the gold proxies on the 60 outdoor test clips from a person F1 of 0.372 to 0.815 (recall 0.231 to 0.750) and a vehicle F1 of 0.420 to 0.478 (\cref{tab:distill}; ground and structure in \cref{tab:distill_full}).
\begin{table}[t]
\caption{Agreement with the gold proxies on the keyframes of the 60 outdoor test clips (center distance $\leq 2$\,m).}
\label{tab:distill}
\centering
\footnotesize
\setlength{\tabcolsep}{3pt}
\begin{tabular*}{\linewidth}{@{\extracolsep{\fill}}lrrrrrr@{}}
\toprule
 & \multicolumn{3}{c}{Vehicle} & \multicolumn{3}{c}{Person} \\
\cmidrule(lr){2-4}\cmidrule(l){5-7}
Scene & P & R & F1 & P & R & F1 \\
\midrule
Perception initialization & 0.605 & 0.322 & 0.420 & 0.953 & 0.231 & 0.372 \\
Automated harness         & 0.567 & 0.414 & 0.478 & 0.892 & 0.750 & 0.815 \\
\bottomrule
\end{tabular*}
\end{table}

\subsection{Video-to-proxy and proxy-to-video baselines}
\label{sec:method:tasks}
\begin{figure*}[t]
\centering
\includegraphics[width=\linewidth]{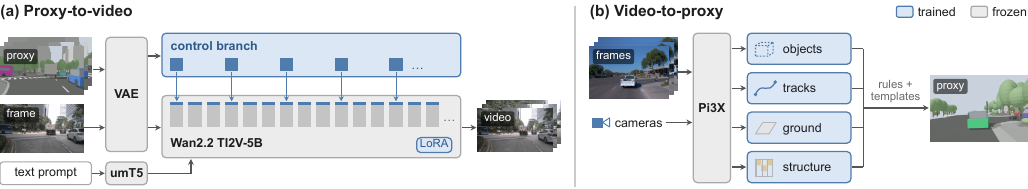}
\caption{Baselines for both directions of the video--proxy interface. (a) Proxy-to-video baseline: a frozen Wan2.2 TI2V-5B generator with a VACE-style control branch and LoRA adapters reads proxy renderings as structured controls. (b) Video-to-proxy baseline: four jointly trained heads on a frozen Pi3X backbone predict objects, tracks, ground and static structure from posed video in one forward pass; rules and templates assemble the outputs into an editable proxy scene.}
\label{fig:arch}
\end{figure*}

ProxyVerse supports both directions of the video--proxy interface.
Our feed-forward video-to-proxy baseline reconstructs an editable proxy from posed video, and our proxy-to-video baseline generates video from a rendered proxy (\cref{fig:arch}).

\paragraph{Video-to-proxy.}
From eight posed frames, a frozen Pi3X backbone~\cite{wang2026pi3} extracts aligned per-frame features and depth.
Four jointly trained heads (25.9\,M parameters) predict 27-class 3D boxes, object trajectories, ground semantics and height, and BEV occupancy of buildings, walls and fences.
Deterministic association links predictions over time, and geometric rules and domain templates assemble one forward pass into the typed representation of \cref{sec:dataset:repr}.
We train driving-only and multi-domain variants on automated proxies (\cref{sec:suppl:train}).

\paragraph{Proxy-to-video.}
We adapt frozen Wan2.2 TI2V-5B~\cite{wan2025wan} with a VACE-style branch~\cite{jiang2025vace} that reads the rendered proxy, together with LoRA adapters on the backbone.
It serves as the proxy-to-video baseline of the benchmark (\cref{sec:exp:p2v}).

\section{Experiments}
\label{sec:exp}
\setcounter{dbltopnumber}{3}
\renewcommand{\dbltopfraction}{0.95}

We organize the experiments around three questions: can a feed-forward model reconstruct an editable proxy directly from video; do proxies provide effective structured control for video generation; and does the representation support controllable editing?

\begin{figure*}[t]
\centering
\includegraphics[width=\linewidth]{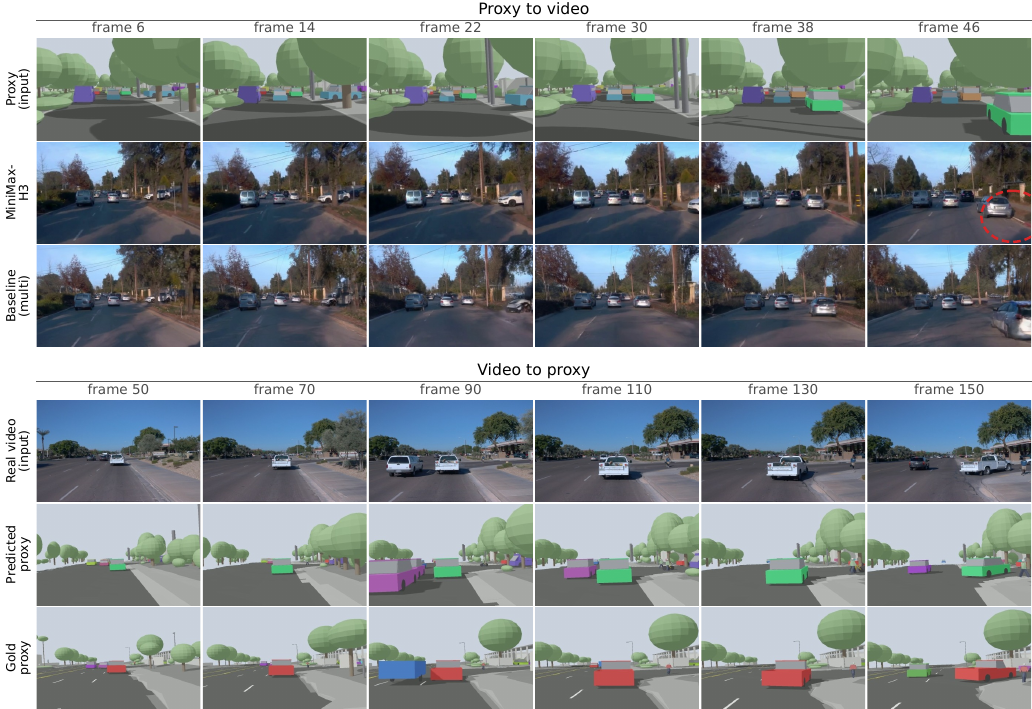}
\caption{Qualitative results on two driving test clips, six frames each. Top: proxy-to-video evaluation, showing the input proxy and videos generated by MiniMax-H3 and our multi-domain baseline. The dashed circle marks where the proxy places an approaching car that the MiniMax-H3 video keeps at a distance. Bottom: video-to-proxy reconstruction, showing the input video, the proxy predicted by our baseline and the gold proxy. Results in all three domains: \cref{fig:p2v_qual,fig:v2p_qual}.}
\label{fig:qual}
\end{figure*}

We evaluate driving-only and multi-domain models on the test split in all three domains. Video-to-proxy models are trained on automated proxies and evaluated against gold proxies; proxy-to-video models are evaluated using gold proxies as control inputs. Training details and metric definitions are given in \cref{sec:suppl:train,sec:suppl:metrics}.

\subsection{Video-to-proxy reconstruction}
\label{sec:exp:v2p}

\paragraph{Setup.}
We evaluate the video-to-proxy baseline of \cref{sec:method:tasks} in driving-only and multi-domain variants (\cref{sec:suppl:train}).
Both variants take 8 posed frames and reconstruct all proxy layers in one forward pass; they differ only in their training domains.
We run DetAny3D~\cite{zhang2025detect}, a released monocular 3D detector run per frame, on driving and outdoor clips, and SpatialLM~1.1~\cite{mao2025spatiallm} and Boxer~\cite{detone2026boxer} on indoor windows, all with their official weights and default settings.

\begin{table*}[t]
\caption{Video-to-proxy reconstruction against gold proxies on whole clips: 185 driving clips, 60 outdoor clips and 309 indoor windows from four homes not seen in training. Our baseline predicts objects, tracks, ground and static structure jointly; released comparison methods cover subsets of these layers. Object F1 uses 3D IoU $\geq 0.25$ in driving and a center distance $\leq 2$\,m elsewhere. ID switches are totals; heights are median errors; mIoU averages road, sidewalk, grass and curb in driving and is floor IoU indoors. BEV: bird's-eye-view occupancy. Indoor per-class F1 and released indoor methods: \cref{tab:v2p_indoor}. N/A: not an output of the method; --: not evaluated. Bold: best in each domain.}
\label{tab:v2p}
\centering
\footnotesize
\setlength{\tabcolsep}{4pt}
\begin{tabular*}{\linewidth}{@{\extracolsep{\fill}}llrrrrrrrrrr@{}}
\toprule
 & & \multicolumn{3}{c}{Object F1$\uparrow$} & \multicolumn{2}{c}{Tracks} & \multicolumn{3}{c}{Ground} & \multicolumn{2}{c}{BEV IoU$\uparrow$} \\
\cmidrule(lr){3-5}\cmidrule(lr){6-7}\cmidrule(lr){8-10}\cmidrule(lr){11-12}
 & Method & All & Vehicle & Person & F1$\uparrow$ & ID sw.$\downarrow$ & Acc.$\uparrow$ & mIoU$\uparrow$ & Height (m)$\downarrow$ & Building & Wall \\
\midrule
\multirow{3}{*}{Driving}
 & DetAny3D              & 0.353 & 0.524 & 0.068 & N/A & N/A & N/A & N/A & N/A & N/A & N/A \\
\cmidrule(l){2-12}
 & Baseline (driving)    & \textbf{0.469} & \textbf{0.640} & \textbf{0.098} & \textbf{0.471} & \textbf{2439} & \textbf{0.970} & \textbf{0.581} & \textbf{0.077} & \textbf{0.247} & 0.113 \\
 & Baseline (multi)      & 0.446 & 0.618 & 0.086 & 0.438 & 3246 & 0.969 & 0.576 & \textbf{0.077} & 0.239 & \textbf{0.119} \\
\midrule
\multirow{2}{*}{Outdoor}
 & DetAny3D              & -- & 0.220 & 0.308 & N/A & N/A & N/A & N/A & N/A & N/A & N/A \\
\cmidrule(l){2-12}
 & Baseline (multi)      & -- & \textbf{0.404} & \textbf{0.609} & 0.359 & 9368 & 0.943 & -- & 0.145 & 0.154 & 0.056 \\
\midrule
Indoor
 & Baseline (multi)      & -- & -- & 0.929 & 0.781 & 151 & 0.998 & 0.994 & 0.030 & -- & 0.382 \\
\bottomrule
\end{tabular*}
\end{table*}

\paragraph{Results.}
\emph{Complete proxy reconstruction.}
Released methods each recover individual scene components, whereas our baseline predicts the major layers needed to assemble an editable proxy (\cref{tab:v2p}).
In driving and outdoors, our baselines have higher object F1 than DetAny3D for every reported class (driving: 0.469 and 0.446 against 0.353), and the multi-domain baseline remains above DetAny3D at every evaluated center-distance threshold (\cref{tab:v2p_thresh}).
\emph{Complementary scene layers.}
Indoors (\cref{tab:v2p_indoor}), our baseline and the released detectors are complementary per class: ours leads on walls, cabinets, doors and windows, has a higher wall BEV IoU than SpatialLM (0.382 against 0.164), and alone predicts people, their tracks and the floor.
\emph{Automated proxies as supervision.}
The baseline is trained on automated proxies and evaluated against gold proxies; it predicts all major proxy layers, showing that the automated proxies of ProxyVerse provide useful supervision for direct video-to-proxy learning (\cref{fig:qual}, bottom).

\subsection{Proxy-to-video generation}
\label{sec:exp:p2v}

\paragraph{Setup.}
Our proxy-to-video baseline adapts the frozen Wan2.2 TI2V-5B model~\cite{wan2025wan} with a VACE-style control branch~\cite{jiang2025vace} and LoRA adapters on the backbone, for 1.82\,B trainable parameters (\cref{fig:arch}a).
It takes the first frame, a text prompt and the proxy rendering, and generates 49 frames at 480p.
We train a driving-only version on the driving training split and a multi-domain version on the driving, outdoor and indoor training splits.
\emph{Wan2.2} is the same backbone given the first frame and prompt only; \emph{Wan2.2-Fun depth} is the public depth-control model of the same family~\cite{alibabapai2025wan22fun5bcontrol}, given the proxy's metric depth; \emph{MiniMax-H3} is a much larger commercial model given the same proxy and first frame.
All methods run on the test clips of \cref{tab:p2v} with gold proxies. The four methods on the Wan2.2 backbone receive the same first frame and prompt and generate one sample per clip with 40 sampling steps and guidance 3.0; MiniMax-H3 runs with its official weights and default settings.
We report PSNR, SSIM and LPIPS against the real frames, camera rotation error and ATE of the generated video, recall of the proxy's movable objects by a detector~\cite{liu2024groundingdino}, and FVD.

\begin{table*}[t]
\caption{Proxy-to-video evaluation on the test sets (49 frames per clip), using gold proxies as structured controls. Driving pools the test clips of PhysicalAI-AV, Waymo and nuScenes; the 309 indoor windows come from four homes not seen in training. Rot.: rotation error in degrees; ATE: relative to trajectory length; Obj.: object recall (the detector recalls 0.64 of the objects in the real videos); FVD is computed over each set. Above the rule: existing models with official weights (MiniMax-H3 is a much larger commercial model given the same proxy); below: our proxy-to-video baselines and, in gray, the multi-domain baseline given an empty proxy (w/o proxy; FVD not computed). Bold: best in each column, commercial model included.}
\label{tab:p2v}
\centering
\footnotesize
\newcommand*{\ab}[1]{{\color{black!55}#1}}
\setlength{\tabcolsep}{0pt}
\begin{tabular*}{\linewidth}{@{\extracolsep{\fill}}l rrrrrrr rrrrrr rrrrrr@{}}
\toprule
 & \multicolumn{7}{c}{Driving (185 clips)} & \multicolumn{6}{c}{Indoor (309 windows)} & \multicolumn{6}{c}{Outdoor (60 clips)} \\
\cmidrule(lr){2-8}\cmidrule(lr){9-14}\cmidrule(l){15-20}
Method & PSNR$\uparrow$ & SSIM$\uparrow$ & LPIPS$\downarrow$ & Rot.$\downarrow$ & ATE$\downarrow$ & Obj.$\uparrow$ & FVD$\downarrow$
 & PSNR$\uparrow$ & SSIM$\uparrow$ & LPIPS$\downarrow$ & Rot.$\downarrow$ & ATE$\downarrow$ & FVD$\downarrow$
 & PSNR$\uparrow$ & SSIM$\uparrow$ & LPIPS$\downarrow$ & Rot.$\downarrow$ & ATE$\downarrow$ & FVD$\downarrow$ \\
\midrule
Wan2.2             & 16.44 & 0.592 & 0.454 & 2.22 & 0.069 & 0.109 & 481.5 & 13.57 & 0.481 & 0.571 & 16.44 & 0.118 & 224.3 & 14.31 & 0.368 & 0.438 & 3.01 & 0.181 & 693.0 \\
Wan2.2-Fun depth   & 17.46 & 0.626 & 0.402 & 1.00 & 0.026 & 0.421 & 271.9 & 15.33 & 0.562 & 0.475 & 2.39 & 0.081 & 224.0 & 16.49 & 0.457 & 0.348 & 0.63 & 0.043 & 469.1 \\
MiniMax-H3         & 20.12 & 0.675 & 0.338 & 0.60 & 0.013 & \textbf{0.451} & 147.0 & 17.62 & 0.622 & 0.381 & 2.70 & 0.079 & 113.7 & 17.28 & 0.461 & 0.310 & 0.46 & 0.042 & 370.9 \\
\midrule
Baseline (driving) & \textbf{21.01} & 0.699 & \textbf{0.287} & \textbf{0.52} & \textbf{0.011} & 0.428 & \textbf{138.5} & 17.88 & 0.634 & 0.361 & 1.88 & 0.082 & 124.4 & 17.73 & \textbf{0.490} & 0.274 & 0.40 & \textbf{0.038} & 332.1 \\
Baseline (multi)   & \textbf{21.01} & \textbf{0.701} & 0.294 & 0.57 & 0.013 & 0.383 & 142.4 & \textbf{18.73} & \textbf{0.675} & \textbf{0.328} & \textbf{1.63} & \textbf{0.071} & \textbf{107.3} & \textbf{17.82} & 0.483 & \textbf{0.269} & \textbf{0.37} & 0.039 & \textbf{288.0} \\
\ab{\quad w/o proxy} & \ab{16.55} & \ab{0.609} & \ab{0.468} & \ab{3.30} & \ab{0.046} & \ab{0.073} & \ab{--} & \ab{13.56} & \ab{0.514} & \ab{0.600} & \ab{40.39} & \ab{0.132} & \ab{--} & \ab{14.32} & \ab{0.373} & \ab{0.450} & \ab{3.57} & \ab{0.046} & \ab{--} \\
\bottomrule
\end{tabular*}
\end{table*}

\begin{figure*}[t]
\centering
\includegraphics[width=\linewidth]{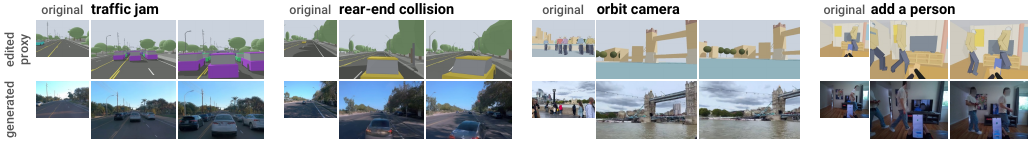}
\caption{Editing a video in proxy space. In each example, the small column shows the original proxy and video, the top row the edited proxy and the bottom row the video regenerated by the multi-domain baseline. From left to right: adding a queue of stopped cars; making the ego car collide with a merging car that brakes; orbiting the camera 30$^\circ$ around a bridge tower; and adding a second person.}
\label{fig:edit}
\end{figure*}

\paragraph{Results.}
\emph{The proxy supplies the control.}
Given an empty proxy (gray row), the multi-domain baseline loses 4.5, 5.2 and 3.5\,dB PSNR on driving, indoors and outdoors, its indoor rotation error rises from 1.63$^\circ$ to 40.39$^\circ$, and its driving object recall falls from 0.383 to 0.073, below Wan2.2 without control (0.109).
\emph{Generation quality.}
Training with ProxyVerse proxies gives consistent gains in every domain (\cref{tab:p2v}, \cref{fig:qual} top): both baselines improve all image metrics and the rotation error over the two references on the same backbone, with PSNR higher by 3.6 to 4.6\,dB on driving, 2.6 to 5.2\,dB indoors and 1.2 to 3.5\,dB outdoors.
\emph{One model across domains.}
One multi-domain baseline covers all three domains: on driving it matches the driving-only baseline in PSNR (21.01\,dB each) and SSIM, with a lower object recall (0.383 against 0.428); it is 0.85\,dB above it indoors, level with it outdoors, and best in every indoor column.
Against the much larger MiniMax-H3 given the same proxy, the multi-domain baseline has higher PSNR in all three domains, by 0.88, 1.11 and 0.55\,dB on driving, indoors and outdoors (95\% paired bootstrap intervals $[0.54, 1.25]$, $[0.92, 1.31]$ and $[0.10, 1.09]$; driving-only on driving: 0.89, $[0.54, 1.28]$), with higher SSIM and lower LPIPS; MiniMax-H3 has the highest object recall on driving.
Indoors and outdoors the multi-domain baseline has the lowest FVD of all methods, the commercial model included (107.3 and 288.0; MiniMax-H3 is second indoors, 113.7); on driving the driving-only baseline has the lowest FVD (138.5), followed by the multi-domain baseline (142.4) and MiniMax-H3 (147.0).
On 45 seen-home windows the multi-domain baseline is again best on every indoor metric, MiniMax-H3 included (\cref{tab:p2v_indoor_full}).
\begin{table}[t]
\caption{Proxy-to-video evaluation of the multi-domain baseline, without further training, on 171 clips with gold proxies from eight external public datasets (Oxford Spires, 3DPW, JRDB, DL3DV, SpatialVID, Dronescapes, EgoBody and ADT). Rot.: median rotation error in degrees; ATE: median, relative to trajectory length, defined on 151 of the 171 clips.}
\label{tab:p2v_xdata}
\centering
\footnotesize
\setlength{\tabcolsep}{2.5pt}
\begin{tabular}{@{}lrrrrr@{}}
\toprule
Method & PSNR$\uparrow$ & SSIM$\uparrow$ & LPIPS$\downarrow$ & Rot.$\downarrow$ & ATE$\downarrow$ \\
\midrule
Wan2.2 & 11.38 & 0.377 & 0.620 & 13.70 & 0.101 \\
Baseline (multi) & \textbf{15.04} & \textbf{0.495} & \textbf{0.431} & \textbf{2.65} & \textbf{0.049} \\
\quad w/o proxy & 11.79 & 0.396 & 0.620 & 23.87 & 0.082 \\
\bottomrule
\end{tabular}
\end{table}

\paragraph{Other datasets.}
\Cref{tab:p2v_xdata} applies the multi-domain baseline, without further training, to 171 clips with gold proxies from eight external public datasets (100 outdoor, 71 indoor); the baseline is 3.65\,dB above Wan2.2 (95\% paired bootstrap interval [3.38, 3.93]) and 3.25\,dB above itself without the proxy ([2.93, 3.60]), with PSNR nearly the same on the outdoor and indoor subsets (15.03 and 15.05\,dB).

\subsection{Proxy editing}
\label{sec:exp:edit}

An edit changes objects, their motion or the camera path of a proxy, and the rendered edited proxy becomes the control input of the multi-domain baseline (\cref{fig:edit,fig:edit_suppl}).
The first frame stays the real one, so the queued cars and the collision in \cref{fig:edit} come from the proxy alone.

\FloatBarrier

\section{Conclusion}
\label{sec:concl}
We introduced ProxyVerse, an open dataset pairing 112.0 hours of real video across driving, outdoor and indoor domains with frame-aligned, editable 3D proxies.
To construct these proxies at scale, human-guided recursive self-improvement consolidates reusable improvements into a harness that open models execute across the remaining clips.
Building on this supervision, we introduced the new video-to-proxy task and a feed-forward baseline that reconstructs the major layers of an editable proxy directly from posed video.
In the reverse direction, proxy-to-video experiments show that proxies provide effective structured control for video generation and editing, supporting them as a bidirectional interface between video understanding and controllable generation.
We will release the videos, proxies, renderings, harness code and baselines for both tasks.

\ifx\acknowledgmentstext\empty\else
\section*{Acknowledgments}
\acknowledgmentstext
\fi
\section*{LLM Usage Statement}
\label{sec:llm}

The research ideas and the method are the authors' own.
We used LLM-based assistants (Claude and GPT models) for four things: to survey related work; to draft text and figures along the direction the authors set; to write code and run experiments, including launching runs in parallel, trying hyperparameter settings and running the compared methods; and to check numbers between tables and text.
The authors inspected and verified everything the assistants produced, including the figures, the drafts and the results, and take full responsibility for the content of the paper.
Separately, large language models are components of the data pipeline itself, as \cref{sec:pipeline} describes: a closed multimodal model (GPT, with feedback from Claude Opus) builds the gold proxies and open models (DeepSeek and Qwen) build and repair the automated proxies.

{
    \small
    \bibliographystyle{ieeenat_fullname}
    \bibliography{references}
}

\clearpage
\twocolumn[{\centering\Large\textbf{Appendix}\par\vspace{1.0em}}]
\renewcommand{\thesection}{S\arabic{section}}
\renewcommand{\thetable}{S\arabic{table}}
\renewcommand{\thefigure}{S\arabic{figure}}
\setcounter{section}{0}
\setcounter{table}{0}
\setcounter{figure}{0}
\renewcommand{\dbltopfraction}{0.95}\renewcommand{\dblfloatpagefraction}{0.85}\renewcommand{\topfraction}{0.9}\renewcommand{\floatpagefraction}{0.85}\renewcommand{\textfraction}{0.05}\setcounter{dbltopnumber}{3}\setcounter{topnumber}{3}\setcounter{totalnumber}{5}

\noindent\textbf{Overview video.} The project page (\projecturl){} begins with a three-minute overview video with sound. It walks through the dataset, the harness-distillation pipeline that builds it, the two baselines (proxy to video and video to proxy), and five uses of the proxies: generating video from a proxy, editing the proxy to obtain a new video, re-shooting a drive from new cameras, recovering an editable proxy from video, and the full loop from video to proxy, edit and back to video. Re-shot views are rendered by MiniMax-H3 from our proxies; all other generated clips come from our Baseline (multi-domain). The rest of the page gives the results section by section, including the re-shoot examples and a comparison with MiniMax-H3 (\cref{sec:suppl:web}).

\section{Real-video datasets with 3D annotation}
\label{sec:suppl:compare_real}
\Cref{tab:compare_real} compares ProxyVerse with existing real-video datasets that provide 3D annotation.
\providecommand{\cmark}{\checkmark}
\providecommand{\xmark}{$\times$}
\providecommand{\nomark}{--}
\providecommand{\pmark}{$\circ$}
\begin{table*}[!t]
\caption{ProxyVerse and existing real-video datasets with 3D annotation. Entries follow the cited papers; scale is quoted in the units of each paper. \emph{Static scene in 3D} and \emph{Dynamic objects in 3D} require object-level 3D annotation (primitives, boxes, CAD models or a labeled mesh; boxes or poses over time for moving objects). \emph{Per-frame rendering}: the paper describes a released 2D rendering of the 3D annotation in the camera view; any row with cameras and 3D labels can be projected by the user. \emph{Automatic annotation}: the camera and 3D annotation are produced without per-clip human labeling or motion capture. \pmark: partial; \nomark: not stated, or no camera or 3D annotation to produce.
$^{a}$Map layers in the ground plane.
$^{b}$Per-pixel or per-point geometry (depth, points, point tracks, 2D masks) without object-level labels.
$^{c}$Automated proxies are built without per-clip human labeling; gold proxies are reviewed frame by frame. Their inputs include the driving datasets' box annotations and indoor motion capture.
$^{d}$In the cited 2020 paper.
$^{e}$2D labels are transferred from the labeled 3D primitives.
$^{f}$Class-agnostic boxes.
$^{g}$From the official repository, not the paper.
$^{h}$For a subset of the trajectories.}
\label{tab:compare_real}
\centering
\small
\setlength{\tabcolsep}{3.5pt}
\begin{tabular}{@{}l l l c c c c c@{}}
\toprule
 & & & Camera & Static scene & Dynamic objects & Per-frame & Automatic \\
Dataset & Domain & Scale & poses & in 3D & in 3D & rendering & annotation \\
\midrule
\multicolumn{8}{@{}l}{\textit{Driving, 3D object annotation}} \\
nuScenes \cite{caesar2020nuscenes} & driving & 1k scenes, 5.5\,h & \cmark & \pmark$^{a}$ & \cmark & \xmark & \xmark \\
Waymo Open \cite{sun2020scalability} & driving & 1{,}150 scenes, 6.4\,h & \cmark & \xmark$^{d}$ & \cmark & \xmark & \xmark \\
Argoverse 2 \cite{wilson2021argoverse2} & driving & 1{,}000 scenes & \cmark & \xmark$^{a}$ & \cmark & \xmark & \xmark \\
KITTI-360 \cite{liao2021kitti360} & driving & 300k images & \cmark & \cmark & \cmark & \cmark$^{e}$ & \xmark \\
\midrule
\multicolumn{8}{@{}l}{\textit{Indoor, 3D objects and layout}} \\
ScanNet++ \cite{yeshwanth2023scannet} & indoor & 460 scenes & \cmark & \cmark & \xmark & \nomark & \xmark \\
CA-1M \cite{lazarow2025cubify} & indoor & 3.5k captures & \cmark & \cmark$^{f}$ & \xmark & \cmark & \xmark \\
CAD-Estate \cite{maninis2023cadestate} & indoor & 19.5k videos & \cmark & \cmark & \xmark & \xmark & \xmark \\
EmbodiedScan \cite{wang2024embodiedscan} & indoor & 5{,}185 scans & \cmark & \cmark & \xmark & \xmark & \xmark \\
ADT \cite{pan2023aria} & indoor, ego. & 200 sequences & \cmark & \cmark & \cmark & \cmark & \xmark \\
\midrule
\multicolumn{8}{@{}l}{\textit{Large-scale video}} \\
MiraData \cite{ju2024miradata} & open & 16k\,h & \xmark & \xmark & \xmark & \xmark & \nomark \\
DL3DV-10K \cite{ling2024dl3dv10k} & in/outdoor & 10{,}510 videos & \cmark$^{g}$ & \xmark & \xmark & \xmark & \cmark \\
Stereo4D \cite{jin2025stereo4d} & open & $>$100k clips & \cmark & \xmark$^{b}$ & \xmark$^{b}$ & \xmark & \cmark \\
Sekai-Real \cite{li2025sekai} & walk, drone & $>$5{,}000\,h & \cmark$^{h}$ & \xmark & \xmark & \xmark & \cmark \\
SpatialVID \cite{wang2026spatialvid} & open & 7{,}089\,h & \cmark & \xmark$^{b}$ & \xmark$^{b}$ & \xmark & \cmark \\
\midrule
ProxyVerse (ours) & driving, walk, indoor ego. & 14{,}923 clips, 112.0\,h & \cmark & \cmark & \cmark & \cmark & \pmark$^{c}$ \\
\bottomrule
\end{tabular}
\end{table*}

\section{Dataset scale}
\label{sec:suppl:scale}
\begin{table}[!htb]
\caption{Scale of ProxyVerse. Hours count non-overlapping video time. Sources are drive logs for driving, source videos for outdoor, and recordings for indoor.}
\label{tab:scale}
\centering
\small
\setlength{\tabcolsep}{4pt}
\begin{tabular}{lrrrr}
\toprule
Domain / subset & Clips & Hours & Frames & Sources \\
\midrule
\multicolumn{5}{l}{\textit{Driving}} \\
\quad PhysicalAI-AV & 1{,}906 & 10.59 & 381k & 1{,}902 \\
\quad Waymo & 957 & 5.27 & 190k & 957 \\
\quad nuScenes & 657 & 3.58 & 153k & 68 \\
\quad all & 3{,}520 & 19.45 & 724k & 2{,}927 \\
\midrule
Outdoor & 2{,}321 & 34.46 & 1{,}802k & 219 \\
Indoor & 9{,}082 & 58.08 & 3{,}136k & 757 \\
\midrule
Total & 14{,}923 & 111.99 & 5{,}663k & 3{,}903 \\
\bottomrule
\end{tabular}
\end{table}

In driving, vegetation accounts for 65\% of the 784k object instances, followed by street furniture (109k), vehicles (94k), pedestrians (27k) and buildings (24k). Outdoors, pedestrians are the second largest class (94k of 367k instances) after buildings (140k).

\section{Perception models}
\label{sec:suppl:percep}
In driving, we fuse depth from Depth Anything~3~\cite{lin2026depth}, VGGT-Omega~[S1], Pi3X~\cite{wang2026pi3,wang2025pi3x} and MoGe-2~[S2] with LiDAR, segment with SegFormer~[S3], Mask2Former~[S4] and SAM~3~\cite{carion2026sam} (prompted by the 3D boxes and by text), and fit boxes to buildings, poles, signs and trees and rules to the ground and lanes.
Outdoors, cameras come from VGGT-Omega; depth is the per-pixel median of Depth Anything~3~\cite{lin2026depth}, VGGT-Omega and Pi3X~\cite{wang2026pi3,wang2025pi3x} after alignment, with metric scale from MoGe-2; Mask2Former gives semantic segmentation, SAM~3~\cite{carion2026sam} instances and their tracks, D-FINE~[S5] with ByteTrack~[S6] the detection boxes (with YOLOv8~[S7] on keyframes for crowds and riders), and Human3R~[S8] the body poses as SMPL-X~\cite{pavlakos2019expressive}.

\section{General review questions}
\label{sec:suppl:general_q}
Asked a general question, whether an edit made a scene better or worse, the open vision-language model we tested caught 17 to 30 of 45 harmful repairs indoors, depending on the prompt, while rejecting 4 to 5 of 23 harmless ones; on a smaller driving set it caught 4 of 6 harmful edits and rejected 2 of 8 good ones.
The harness therefore uses narrow questions with a few possible answers instead (\cref{sec:pipeline:hd}).

\section{Training details}
\label{sec:suppl:train}

\paragraph{Proxy-to-video.}
The baseline builds on the Wan2.2 TI2V-5B model~\cite{wan2025wan}, whose transformer, VAE and text encoder stay frozen.
The proxy rendering is encoded by the same VAE and fed to a control branch in the style of VACE~\cite{jiang2025vace}: trainable copies of every third transformer block, whose outputs pass through zero-initialized projections and are added to the frozen blocks as hints.
Rank-32 LoRA adapters are attached to the attention and feed-forward layers of all backbone blocks, which gives 1.82\,B trainable parameters (1.74\,B in the branch, 0.08\,B in the adapters).
The model takes the first frame, a text prompt and the proxy rendering, and generates 49 frames at 480p; it is trained with a flow-matching loss, and the proxy input is dropped for 5\% of the samples.
We train with AdamW on 8 GPUs, with a learning rate of $5\times10^{-5}$ for the branch and $10^{-4}$ for the adapters.
The driving-only version is trained on the driving training split, and the multi-domain version on the driving, outdoor and indoor training splits.
At inference, all methods on the Wan2.2 backbone use 40 sampling steps and a classifier-free guidance scale of 3.0 on the text condition.

\paragraph{Video-to-proxy.}
The center head predicts center heatmaps for 27 classes with 3D box regression, and the track-query head uses its peaks as queries to output the center, heading and visibility of each object over the 8 frames, with its size and score.
The ground head predicts 6 ground classes and the ground height relative to the camera, and the bird's-eye-view head predicts the occupancy and height of buildings, walls and fences.
One set of heads and one class table serve all three domains, with the classes that a domain does not label masked out.
The driving-only version is trained on the driving training split.
The multi-domain version is trained on the training splits of all three domains, followed by a second stage in which the outdoor data is restricted to a subset selected by a camera-motion consistency check.
Outdoors it uses the 1{,}450 training clips whose source video supplies no evaluation clip, so training and test clips never share a source video.

For video-to-proxy, DetAny3D uses the same camera intrinsics as our baseline (the calibrated intrinsics in driving, the reference camera outdoors).

\begin{table*}[!t]
\caption{Proxy-to-video on indoor test windows from homes that also appear in training (seen) and from homes that do not (unseen; the indoor set of \cref{tab:p2v}). Columns as in \cref{tab:p2v}; higher is better for PSNR and SSIM, lower is better for LPIPS, rotation error (degrees) and ATE.}
\label{tab:p2v_indoor_full}
\centering
\footnotesize
\setlength{\tabcolsep}{0pt}
\begin{tabular*}{0.8\linewidth}{@{\extracolsep{\fill}}lrrrrr rrrrr@{}}
\toprule
 & \multicolumn{5}{c}{Seen homes (45 windows)} & \multicolumn{5}{c}{Unseen homes (309 windows)} \\
\cmidrule(lr){2-6}\cmidrule(l){7-11}
Method & PSNR & SSIM & LPIPS & Rot. & ATE & PSNR & SSIM & LPIPS & Rot. & ATE \\
\midrule
Wan2.2             & 13.64 & 0.502 & 0.580 & 16.34 & 0.122 & 13.57 & 0.481 & 0.571 & 16.44 & 0.118 \\
Wan2.2-Fun depth   & 15.03 & 0.569 & 0.499 & 2.59 & 0.081 & 15.33 & 0.562 & 0.475 & 2.39 & 0.081 \\
MiniMax-H3         & 17.37 & 0.619 & 0.404 & 2.54 & 0.075 & 17.62 & 0.622 & 0.381 & 2.70 & 0.079 \\
\midrule
Baseline (driving) & 17.90 & 0.641 & 0.371 & 3.13 & 0.083 & 17.88 & 0.634 & 0.361 & 1.88 & 0.082 \\
Baseline (multi)   & \textbf{19.07} & \textbf{0.682} & \textbf{0.337} & \textbf{1.90} & \textbf{0.070} & \textbf{18.73} & \textbf{0.675} & \textbf{0.328} & \textbf{1.63} & \textbf{0.071} \\
\bottomrule
\end{tabular*}
\end{table*}
\begin{table*}[!t]
\caption{FVD$\downarrow$ with one and three 16-frame windows per clip (frames 1, 17 and 33 for three windows). One window is the protocol of \cref{tab:p2v}. --: not evaluated. Bold: lowest in each column.}
\label{tab:p2v_fvd_windows}
\centering
\footnotesize
\setlength{\tabcolsep}{0pt}
\begin{tabular*}{\linewidth}{@{\extracolsep{\fill}}l rr rr rr rr@{}}
\toprule
 & \multicolumn{2}{c}{Driving 185} & \multicolumn{2}{c}{Indoor seen 45} & \multicolumn{2}{c}{Indoor unseen 309} & \multicolumn{2}{c}{Outdoor 60} \\
\cmidrule(lr){2-3}\cmidrule(lr){4-5}\cmidrule(lr){6-7}\cmidrule(l){8-9}
Method & 1 win. & 3 win. & 1 win. & 3 win. & 1 win. & 3 win. & 1 win. & 3 win. \\
\midrule
Wan2.2             & 481.5 & 415.1 & 670.8 & 430.4 & 224.3 & 158.1 & 693.0 & 561.9 \\
Wan2.2-Fun depth   & 271.9 & 212.8 & 642.7 & 425.9 & 224.0 & 172.1 & 469.1 & 369.7 \\
MiniMax-H3         & 147.0 & -- & 424.2 & 274.9 & 113.7 & 67.7 & 370.9 & 241.5 \\
\midrule
Baseline (driving) & \textbf{138.5} & \textbf{89.8} & 436.7 & 284.1 & 124.4 & 88.8 & 332.1 & 243.2 \\
Baseline (multi)   & 142.4 & 97.0 & \textbf{404.8} & \textbf{262.8} & \textbf{107.3} & \textbf{64.0} & \textbf{288.0} & \textbf{217.0} \\
\bottomrule
\end{tabular*}
\end{table*}
\section{Metric definitions}
\label{sec:suppl:metrics}

\paragraph{Proxy-to-video.}
PSNR, SSIM and LPIPS compare generated and real frames.
Camera error compares the camera trajectory estimated from the generated video with the reference camera of the clip: ATE is the translation error after similarity alignment, relative to the trajectory length, and rotation error is the median over clips of the mean per-frame error in degrees.
Object recall measures whether the generated video puts the movable objects of the proxy where the proxy places them: we project the visible movable objects of the proxy (vehicles, two-wheelers, pedestrians, traffic cones and barriers) into each frame and count a proxy box as recalled if an open-vocabulary detector (Grounding DINO~\cite{liu2024groundingdino}) finds a box of the same class with 2D IoU of at least 0.5 in the generated frame.
FVD uses I3D features of 16 frames per clip over all clips of each test set.

\paragraph{Video-to-proxy.}
Per-frame object F1 matches predicted and gold boxes of the same class, at 3D IoU of at least 0.25 in driving and at a center distance of at most 2\,m outdoors and indoors, so values are not comparable across domains.
Track F1 is a trajectory-level F1 at the same 2\,m center distance, and ID switches are counted over all test clips.
For the ground we report the accuracy of the per-cell ground class, the mean IoU over ground classes (road, sidewalk, grass and curb in driving), or the IoU of the floor indoors, and the median error of the ground height.
Static structure is scored as bird's-eye-view occupancy IoU of buildings and walls.

\section{Proxy-to-video on the full test sets}
\label{sec:suppl:p2v_full}
\Cref{tab:p2v_indoor_full} evaluates proxy-to-video on indoor test windows from seen and unseen homes, with the protocol of \cref{tab:p2v}.

\Cref{tab:p2v_fvd_windows} repeats the FVD evaluation with three windows per clip: both baselines stay below the two same-backbone references in every set, and the multi-domain baseline keeps the lowest FVD in every indoor and outdoor set. FVD decreases with the number of clips, so values are comparable only within a set.

\section{Video-to-proxy: indoor classes}
\label{sec:suppl:v2p_indoor}
\Cref{tab:v2p_indoor} reports video-to-proxy reconstruction on the 309 indoor test windows from homes not seen in training, with object F1 for each indoor class.
\begin{table*}[!t]
\caption{Video-to-proxy reconstruction on the 309 indoor test windows from four homes not seen in training, with per-class object F1 at a center distance $\leq 2$\,m. Our model predicts all listed proxy layers jointly. Columns after the object classes repeat the indoor row of \cref{tab:v2p}; the released comparison methods also receive the device poses and semi-dense points of the recordings. SpatialLM gives no output on one window (counted as empty). N/A: the method does not output this component; --: not evaluated. Bold: best in each column.}
\label{tab:v2p_indoor}
\centering
\footnotesize
\setlength{\tabcolsep}{2.5pt}
\begin{tabular*}{\linewidth}{@{\extracolsep{\fill}}lrrrrrrrrrrrrrrr@{}}
\toprule
 & \multicolumn{9}{c}{Object F1$\uparrow$} & \multicolumn{2}{c}{Person tracks} & \multicolumn{3}{c}{Ground} & BEV IoU$\uparrow$ \\
\cmidrule(lr){2-10}\cmidrule(lr){11-12}\cmidrule(lr){13-15}\cmidrule(lr){16-16}
Indoor & Person & Wall & Cabinet & Seat & Table & Bed & Door & Window & Appl. & F1$\uparrow$ & ID sw.$\downarrow$ & Acc.$\uparrow$ & Floor$\uparrow$ & Height$\downarrow$ & Wall \\
\midrule
SpatialLM 1.1         & -- & 0.383 & 0.122 & 0.202 & 0.185 & 0.571 & 0.071 & 0.063 & 0.049 & N/A & N/A & N/A & N/A & N/A & 0.164 \\
Boxer                 & -- & N/A & 0.345 & \textbf{0.660} & \textbf{0.636} & \textbf{0.691} & 0.312 & 0.226 & \textbf{0.607} & N/A & N/A & N/A & N/A & N/A & N/A \\
\midrule
Baseline (multi) & \textbf{0.929} & \textbf{0.682} & \textbf{0.649} & 0.647 & 0.506 & 0.184 & \textbf{0.515} & \textbf{0.517} & 0.537 & 0.781 & 151 & 0.998 & 0.994 & 0.030 & \textbf{0.382} \\
\bottomrule
\end{tabular*}
\end{table*}

\section{Video-to-proxy: matching thresholds}
\label{sec:suppl:v2p_thresh}
\Cref{tab:v2p_thresh} varies the center-distance threshold of the object match; the baseline leads DetAny3D at every threshold.
\begin{table*}[!t]
\caption{Video-to-proxy object F1 at center-distance thresholds of 0.5, 1 and 2\,m, per frame, on the 185 driving clips and 60 outdoor clips of \cref{tab:v2p} and its 309 indoor windows (driving in \cref{tab:v2p} uses 3D IoU $\geq 0.25$). Bold: best per class and threshold.}
\label{tab:v2p_thresh}
\centering
\footnotesize
\setlength{\tabcolsep}{4pt}
\begin{tabular*}{\linewidth}{@{\extracolsep{\fill}}lllrrr@{}}
\toprule
Domain & Class & Method & 0.5\,m & 1\,m & 2\,m \\
\midrule
\multirow{6}{*}{Driving} & \multirow{2}{*}{All} & DetAny3D & 0.167 & 0.314 & 0.469 \\
 & & Baseline (multi) & \textbf{0.223} & \textbf{0.399} & \textbf{0.573} \\
 & \multirow{2}{*}{Vehicle} & DetAny3D & 0.182 & 0.353 & 0.540 \\
 & & Baseline (multi) & \textbf{0.235} & \textbf{0.422} & \textbf{0.614} \\
 & \multirow{2}{*}{Person} & DetAny3D & 0.176 & 0.320 & 0.464 \\
 & & Baseline (multi) & \textbf{0.254} & \textbf{0.427} & \textbf{0.580} \\
\midrule
\multirow{4}{*}{Outdoor} & \multirow{2}{*}{Vehicle} & DetAny3D & 0.037 & 0.101 & 0.220 \\
 & & Baseline (multi) & \textbf{0.089} & \textbf{0.221} & \textbf{0.404} \\
 & \multirow{2}{*}{Person} & DetAny3D & 0.049 & 0.145 & 0.308 \\
 & & Baseline (multi) & \textbf{0.233} & \textbf{0.424} & \textbf{0.609} \\
\midrule
Indoor & Person & Baseline (multi) & 0.882 & 0.931 & 0.929 \\
\bottomrule
\end{tabular*}
\end{table*}

\section{Distillation fidelity}
\label{sec:suppl:distill}
\Cref{tab:distill_full} extends \cref{tab:distill} with the ground and static structure, and adds the video-to-proxy baseline and DetAny3D scored on the same keyframes.
\begin{table*}[!t]
\caption{Agreement with the gold proxies on the keyframes (1.5 frames per second) of the 60 outdoor test clips. The perception initialization is the starting point of both the gold stage and the automated harness. Objects at a center distance $\leq 2$\,m; ground-class accuracy, per-class IoU and median height error; bird's-eye-view occupancy IoU of static structure. N/A: not an output of the method. All rows are scored on keyframes.}
\label{tab:distill_full}
\centering
\footnotesize
\setlength{\tabcolsep}{0pt}
\begin{tabular*}{\linewidth}{@{\extracolsep{\fill}}l rrr rrr rrrrr rrr@{}}
\toprule
 & \multicolumn{3}{c}{Vehicle} & \multicolumn{3}{c}{Person} & \multicolumn{5}{c}{Ground} & \multicolumn{3}{c}{BEV IoU$\uparrow$} \\
\cmidrule(lr){2-4}\cmidrule(lr){5-7}\cmidrule(lr){8-12}\cmidrule(l){13-15}
Scene & P$\uparrow$ & R$\uparrow$ & F1$\uparrow$ & P$\uparrow$ & R$\uparrow$ & F1$\uparrow$ & Acc.$\uparrow$ & Road$\uparrow$ & Sidewalk$\uparrow$ & Grass$\uparrow$ & Height (m)$\downarrow$ & Building & Wall & Fence \\
\midrule
Perception initialization & 0.605 & 0.322 & 0.420 & 0.953 & 0.231 & 0.372 & 0.940 & 0.561 & 0.764 & 0.466 & 0.047 & 0.141 & 0.106 & 0.048 \\
Automated harness (open models) & 0.567 & 0.414 & 0.478 & 0.892 & 0.750 & 0.815 & 0.935 & 0.564 & 0.681 & 0.638 & 0.050 & 0.165 & 0.037 & 0.117 \\
\midrule
Video-to-proxy baseline (multi) & 0.388 & 0.417 & 0.402 & 0.638 & 0.589 & 0.613 & 0.942 & 0.561 & 0.684 & 0.606 & 0.145 & 0.154 & 0.056 & 0.048 \\
DetAny3D & 0.214 & 0.219 & 0.217 & 0.345 & 0.279 & 0.309 & N/A & N/A & N/A & N/A & N/A & N/A & N/A & N/A \\
\bottomrule
\end{tabular*}
\end{table*}

\section{Proxy-to-video examples}
\label{sec:suppl:p2v_qual}
\begin{figure*}[!t]
\centering
\includegraphics[width=\linewidth]{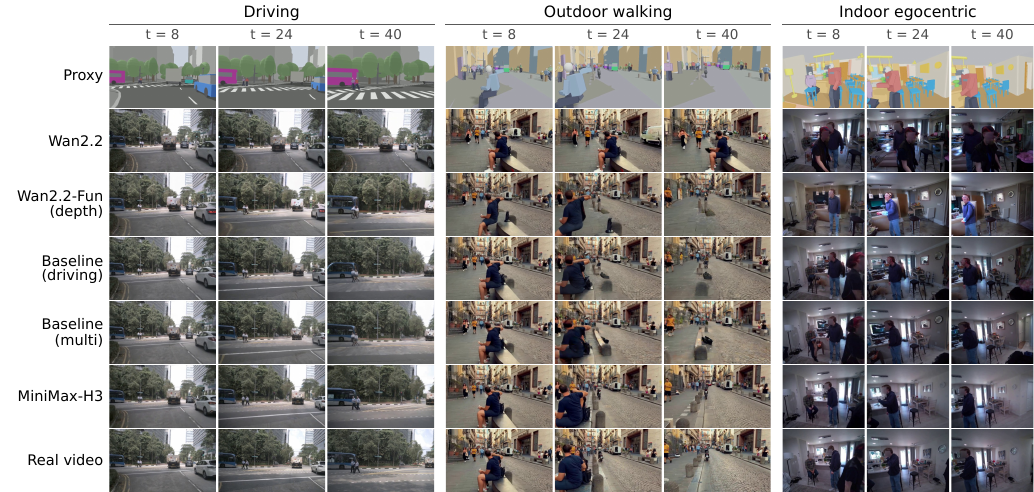}
\caption{Proxy-to-video evaluation on one test clip per domain. Time runs from left to right (frames 8, 24 and 40 of 49); the top row is the proxy rendering used as control, the middle rows are outputs of existing generators and our proxy-to-video baselines, and the bottom row is the real video. The comparison shows how each generator places the vehicles, people and walls specified by the proxy and follows its camera motion.}
\label{fig:p2v_qual}
\end{figure*}

\Cref{fig:p2v_qual} compares the proxy-to-video baselines with existing generators on one test clip per domain.

\section{Video-to-proxy examples}
\label{sec:suppl:v2p_qual}
\begin{figure*}[!t]
\centering
\includegraphics[width=\linewidth]{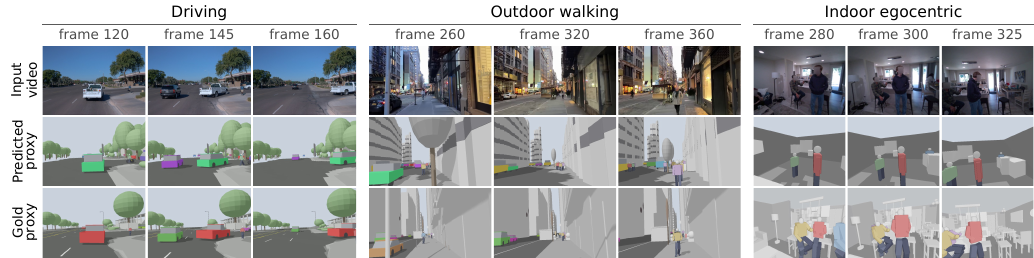}
\caption{Video-to-proxy reconstruction on one test clip per domain; time runs from left to right. Rows: input video, proxy predicted by our feed-forward model in one pass, and gold proxy. The examples show tracked people and vehicles together with recovered ground and walls.}
\label{fig:v2p_qual}
\end{figure*}

\Cref{fig:v2p_qual} shows video-to-proxy reconstructions on one test clip per domain.

\section{More editing examples}
\label{sec:suppl:edit}
\Cref{fig:edit_suppl} shows further edits in the three domains, regenerated by the multi-domain baseline: changed vehicle motion and added cars in driving, new camera paths outdoors, and redirected or added people indoors.

\section{Project page}\label{sec:suppl:web}
The project page (\projecturl) shows{} videos of dataset examples from the three domains, each with its proxy and a 3D view, and follows one driving clip through harness distillation, from perception outputs and the initial proxy to the gold and automated proxies. It also shows the two baselines with their video inputs and outputs. Its results section presents video results for proxy to video, proxy editing, closing the loop and video to proxy, together with further qualitative examples.

\clearpage
\twocolumn[{\centering
\includegraphics[width=\linewidth]{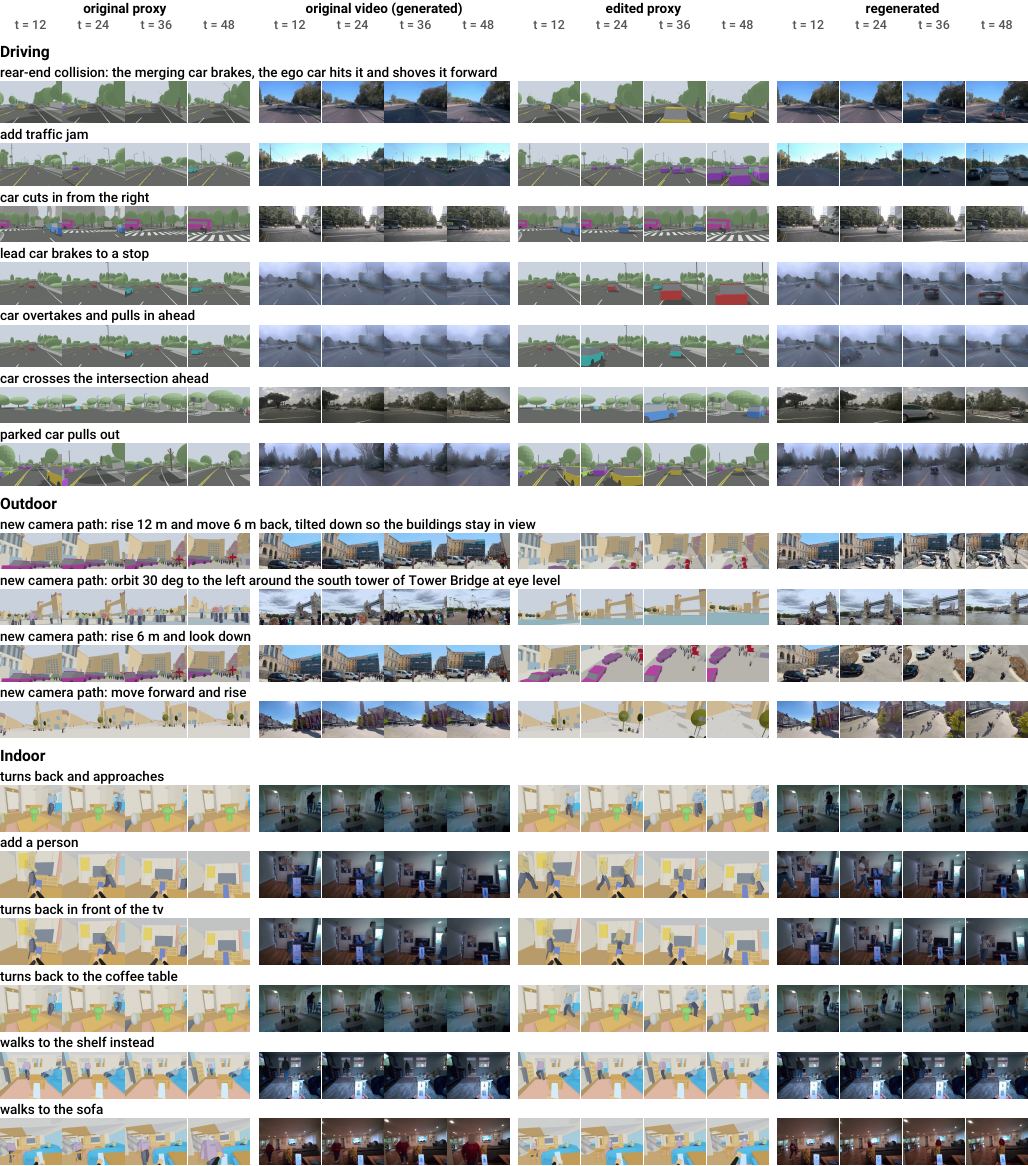}
\captionof{figure}{Proxy-space editing examples. Each row shows the original proxy, the video produced by the proxy-conditioned generator, the edited proxy and the video regenerated from it, at four times. The first frame is the real one in every row. The outdoor camera orbit around a bridge tower is also shown in \cref{fig:edit}.}
\label{fig:edit_suppl}
\vspace{1em}}]

\clearpage
\section*{Additional references}
{\small
\begin{list}{}{\setlength{\leftmargin}{2.4em}\setlength{\labelwidth}{2.1em}\setlength{\labelsep}{0.3em}\setlength{\itemsep}{1pt}\setlength{\parsep}{0pt}\setlength{\topsep}{2pt}}
\item[{[S1]}] Jianyuan Wang, Minghao Chen, Shangzhan Zhang, Nikita Karaev, Johannes Sch{\"o}nberger, Patrick Labatut, Piotr Bojanowski, David Novotny, Andrea Vedaldi, and Christian Rupprecht.
VGGT-$\Omega$.
In \emph{Proceedings of the IEEE/CVF Conference on Computer Vision and Pattern Recognition (CVPR)}, 2026.
\item[{[S2]}] Ruicheng Wang, Sicheng Xu, Yue Dong, Yu Deng, Jianfeng Xiang, Zelong Lv, Guangzhong Sun, Xin Tong, and Jiaolong Yang.
MoGe-2: Accurate Monocular Geometry with Metric Scale and Sharp Details.
In \emph{Advances in Neural Information Processing Systems (NeurIPS)}, 2025.
\item[{[S3]}] Enze Xie, Wenhai Wang, Zhiding Yu, Anima Anandkumar, Jose M. Alvarez, and Ping Luo.
SegFormer: Simple and Efficient Design for Semantic Segmentation with Transformers.
In \emph{Advances in Neural Information Processing Systems (NeurIPS)}, 2021.
\item[{[S4]}] Bowen Cheng, Ishan Misra, Alexander G. Schwing, Alexander Kirillov, and Rohit Girdhar.
Masked-attention Mask Transformer for Universal Image Segmentation.
In \emph{Proceedings of the IEEE/CVF Conference on Computer Vision and Pattern Recognition (CVPR)}, 2022.
\item[{[S5]}] Yansong Peng, Hebei Li, Peixi Wu, Yueyi Zhang, Xiaoyan Sun, and Feng Wu.
D-FINE: Redefine Regression Task of DETRs as Fine-grained Distribution Refinement.
In \emph{International Conference on Learning Representations (ICLR)}, 2025.
\item[{[S6]}] Yifu Zhang, Peize Sun, Yi Jiang, Dongdong Yu, Fucheng Weng, Zehuan Yuan, Ping Luo, Wenyu Liu, and Xinggang Wang.
ByteTrack: Multi-Object Tracking by Associating Every Detection Box.
In \emph{European Conference on Computer Vision (ECCV)}, 2022.
\item[{[S7]}] Glenn Jocher, Ayush Chaurasia, and Jing Qiu.
Ultralytics YOLOv8.
Software, version 8.0.0, 2023.
\item[{[S8]}] Yue Chen, Xingyu Chen, Yuxuan Xue, Anpei Chen, Yuliang Xiu, and Gerard Pons-Moll.
Human3R: Everyone Everywhere All at Once.
In \emph{International Conference on Learning Representations (ICLR)}, 2026.
\end{list}
}

\end{document}